\documentclass[lettersize,journal]{IEEEtran}
\usepackage{amsmath,amsfonts}
\usepackage{algorithmic}
\usepackage{algorithm}
\usepackage{array}
\usepackage[caption=false,font=normalsize,labelfont=sf,textfont=sf]{subfig}
\usepackage{textcomp}
\usepackage{stfloats}
\usepackage{url}
\usepackage{pifont}
\usepackage{verbatim}
\usepackage{graphicx}
\usepackage{braket}
\usepackage{cite}
\usepackage{hyperref}
\hypersetup{
    colorlinks=true,
    linkcolor=blue,
    filecolor=magenta,      
    urlcolor=cyan,	
    pdfpagemode=UseNone,
    citecolor=blue,
    }
\usepackage{booktabs}
\usepackage[table]{xcolor}
\usepackage{multirow}
\usepackage{multicol}
\usepackage{tikz}

\begin{document}

\title{Voronoi-guided Bilateral 2D Gaussian Splatting for Arbitrary-Scale Hyperspectral Image Super-Resolution}

\author{Jie Zhang, Jinkun You,~\IEEEmembership{Graduate Student Member, IEEE}, Shi Chen, Yongshan Zhang,~\IEEEmembership{Senior Member, IEEE}\\ and Yicong Zhou,~\IEEEmembership{Senior Member, IEEE}
        
\thanks{This work was funded in part by the Science and Technology Development Fund, Macau SAR (File no. 0050/2024/AGJ), by the University of Macau and University of Macau Development Foundation (File no. MYRG-GRG2024-00181-FST-UMDF). (Corresponding author: Yicong Zhou.)}
\thanks{Jie Zhang, Jinkun You, Shi Chen and Yicong Zhou are with the Department of Artificial Intelligence, University of Macau, Macau 999078, China (e-mail: jiezh1997@gmail.com; youjinkun09@gmail.com; 
chenshi@um.edu.mo; yicongzhou@um.edu.mo).}
\thanks{Yongshan Zhang is with the School of Software Engineering, Sun Yat-sen University, 519082 China. (e-mail: yszhang.cug@gmail.com)}}

\markboth{submitted}{Shell \MakeLowercase{\textit{et al.}}: A Sample Article Using IEEEtran.cls for IEEE Journals}

\maketitle

\begin{abstract}

Most existing hyperspectral image super-resolution methods require modifications for different scales, limiting their flexibility in arbitrary-scale reconstruction. 2D Gaussian splatting provides a continuous spatial representation that is compatible with arbitrary-scale super-resolution. Existing Gaussian-splatting-based methods often employ rasterization strategies with predefined spatial support regions, which may limit the flexibility of pixel–Gaussian associations. Extending them to hyperspectral image super-resolution remains challenging, as the task requires adaptive spatial reconstruction while preserving spectral fidelity. This paper proposes GaussianHSI, a Gaussian-Splatting-based framework for arbitrary-scale hyperspectral image super-resolution. We develop a Voronoi-Guided Bilateral 2D Gaussian Splatting (VBGS) module for spatial reconstruction. After predicting a set of Gaussian functions to represent the spatial structures of the low-resolution input, it associates each target high-resolution pixel with relevant Gaussian functions through Voronoi-guided selection. The target pixel feature is then reconstructed by aggregating the selected Gaussian functions with reference-aware bilateral weighting, which considers both geometric relevance and consistency with low-resolution features. To complement the spatial structure modeling of VBGS, we further introduce a Spectral Detail Enhancement module to improve spectral reconstruction. Extensive experiments on benchmark datasets demonstrate improved reconstruction performance of our GaussianHSI over the compared state-of-the-art methods for both single-scale and arbitrary-scale hyperspectral image super-resolution.

\end{abstract}

\begin{IEEEkeywords}
Hyperspectral image, arbitrary-scale super-resolution, Gaussian splatting, low-level vision.
\end{IEEEkeywords}

\section{Introduction}
\IEEEPARstart{H}{yperspectral} images (HSIs) capture scene information across numerous narrow spectral bands, providing richer spectral information than RGB images~\cite{hsi2}. They are widely used in remote sensing applications, such as environmental monitoring~\cite{yn} and scene understanding~\cite{ys,fc}. However, due to the trade-off between spatial and spectral resolution in hyperspectral imaging, HSIs often suffer from limited spatial resolution~\cite{hsi,mm3,jxsr}. This limitation may affect their effectiveness in remote sensing tasks. HSI super-resolution aims to reconstruct a high-resolution HSI from a low-resolution observation, enhancing spatial details while preserving spectral fidelity \cite{srreview,denannan}. Therefore, it is important for downstream remote sensing applications.

\begin{figure}[!t]
    \centering
    \includegraphics[width=1\linewidth]{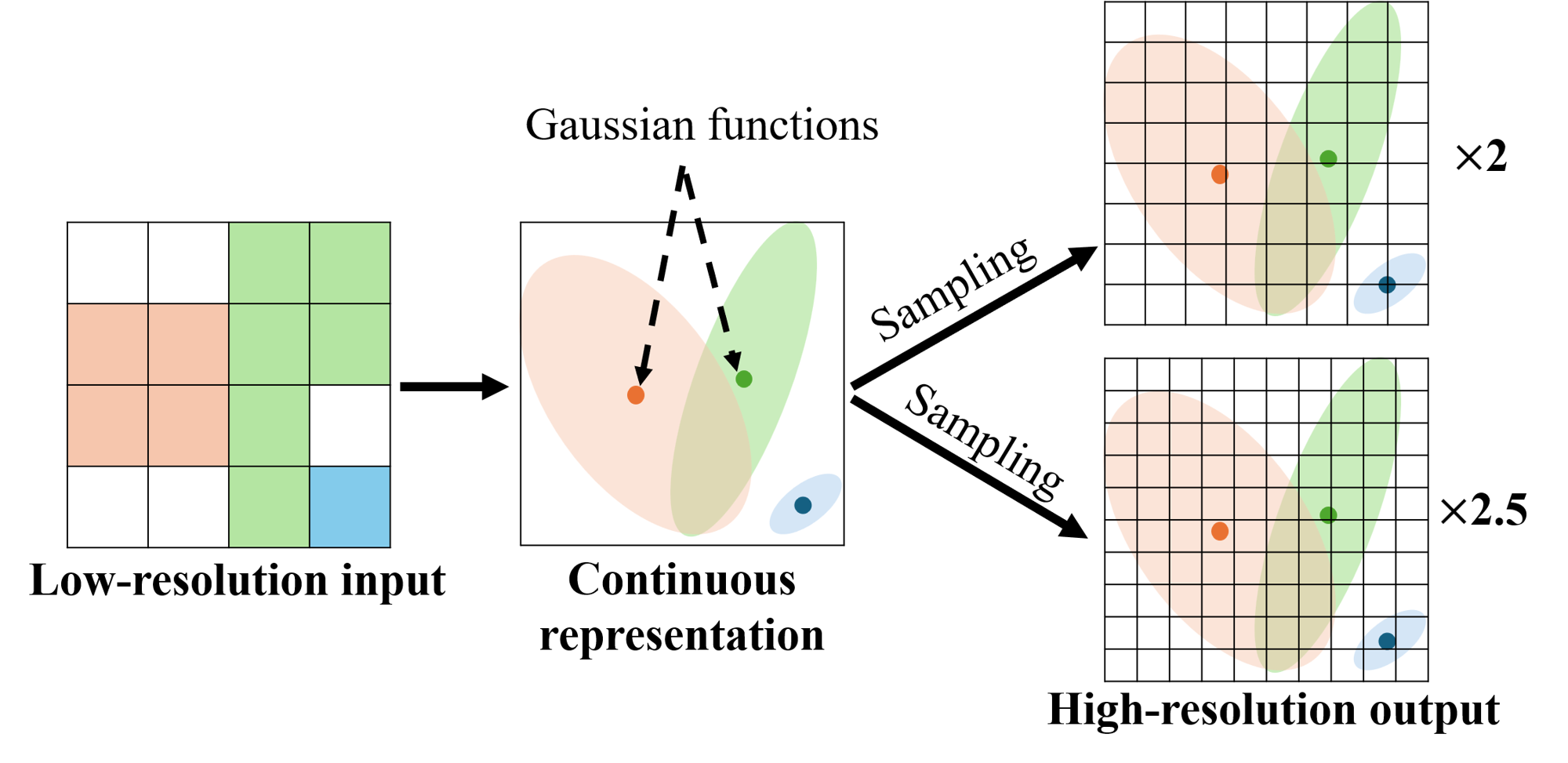}
    \caption{Conceptual illustration of Gaussian splatting for arbitrary-scale image super-resolution. A continuous representation composed of Gaussian functions is inferred from the low-resolution input and rendered on target grids at different scales (e.g., $\times$2 and $\times$2.5). This representation enables arbitrary-scale reconstruction.}
    \label{fig:continue}
\end{figure}

\IEEEpubidadjcol

Deep learning has substantially advanced HSI super-resolution in recent years~\cite{yxf,mcnet,msdformer,cnnzuo}. CNN-based methods have been widely used in HSI super-resolution due to their effectiveness in extracting local features for reconstruction~\cite{mcnet,fy,sspsr}. To capture long-range dependencies, Transformer-based methods have been introduced to model interactions among spatially distant regions~\cite{essa,msdformer,xiaoyi}. More recently, Mamba-based~\cite{mamba} architectures have attracted increasing attention as an efficient alternative to Transformers, showing potential for balancing reconstruction performance and computational complexity~\cite{hsrmamba}. However, most existing HSI super-resolution methods couple network architectures with fixed-scale spatial upsampling operations. As a result, the target scale is typically predefined during model design and training, and the resulting model is usually restricted to that scale. In practical applications, the desired spatial resolution may vary, requiring reconstruction at different scales. These fixed-scale architectures limit model flexibility and often require architectural modifications and separately trained models for different target scales, increasing training costs and storage requirements~\cite{sqformer,gaussiansr}. These limitations motivate the development of arbitrary-scale HSI super-resolution methods that can accommodate different target scales within a single model.

\begin{figure*}
    \centering
    \includegraphics[width=0.99\linewidth]{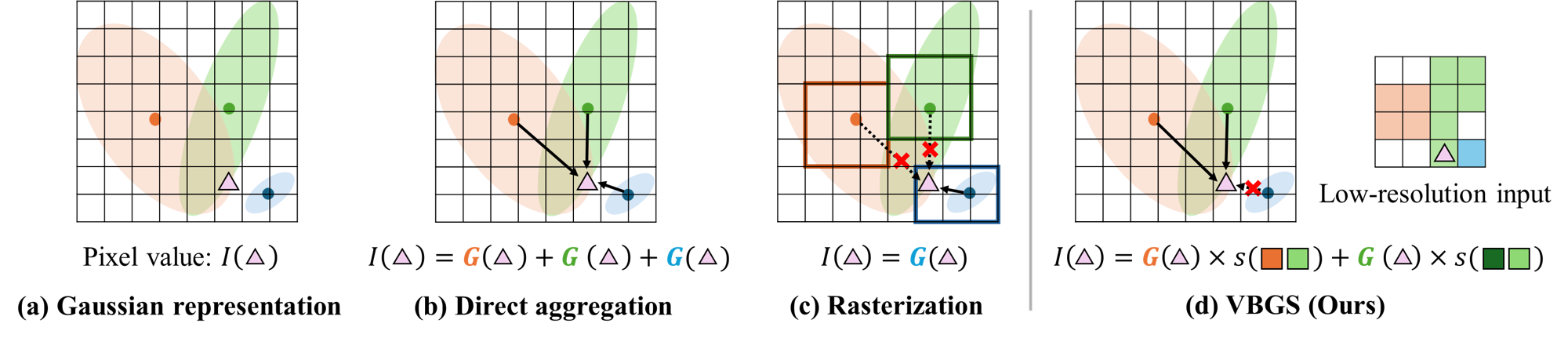}
    \caption{Comparison of different Gaussian aggregation strategies for pixel reconstruction. (a) Gaussian functions provide a continuous image representation.
(b) Direct aggregation computes the target pixel by aggregating all Gaussian functions. (c) Rasterization improves efficiency by restricting aggregation at each target pixel to a subset of nearby Gaussian functions. (d) Voronoi-guided Bilateral 2D Gaussian Splatting (VBGS) performs adaptive aggregation by identifying relevant Gaussian functions for each target pixel and modulating their contributions using normalized reference-aware weights. These weights are computed from the similarities between the selected Gaussian features and the low-resolution reference feature. Here, $G(\triangle)$ denotes the contribution of a Gaussian function at the target pixel, and $s(\cdot,\cdot)$ denotes the similarity.}
    \label{fig:compare}
\end{figure*}

Recent studies have explored arbitrary-scale HSI super-resolution through different strategies~\cite{sqformer,mcarb,msanet}. For example, SQformer~\cite{sqformer} constructs a query token for each target high-resolution pixel and retrieves features from low-resolution tokens to reconstruct that pixel. MCArb~\cite{mcarb} predicts convolutional kernels for different target scales and uses them to upsample low-resolution features. While these methods support reconstruction at different scales, they focus on directly predicting pixel features at the discrete target spatial locations rather than representing spatial structures with continuous representations. As illustrated in Fig.~\ref{fig:continue}, Gaussian functions provide a flexible spatial representation of image content~\cite{gaussianimage,gsasr}. These Gaussian functions can be evaluated at target pixel locations of different resolutions, making this representation suitable for arbitrary-scale super-resolution~\cite{gaussiansr}.

Inspired by the success of Gaussian Splatting in continuous scene representation and efficient rendering~\cite{3dgs}, recent studies have extended Gaussian Splatting to 2D image modeling~\cite{igs,gaussianimage,gsasr,gaussiansr,continuoussr}. By representing image content with Gaussian functions and aggregating them at target locations, 2D Gaussian Splatting provides a continuous reconstruction formulation that is attractive for arbitrary-scale super-resolution~\cite{gaussianimage}. However, directly learning effective Gaussian representations from an input image remains challenging~\cite{continuoussr,igs}. To address this issue, recent studies improve 2D Gaussian modeling through direct pixel-to-Gaussian construction~\cite{gaussiansr}, better Gaussian attribute initialization~\cite{igs}, and prior-guided parameter learning~\cite{continuoussr}. Meanwhile, the rasterization strategy is adopted to improve the efficiency of Gaussian rendering~\cite{gsasr}. These advances suggest the promise of 2D Gaussian Splatting for arbitrary-scale image super-resolution.

Existing Gaussian-based super-resolution methods for color images face challenges when directly applied to HSI reconstruction because of the domain gap between color images and HSIs~\cite{domaingap}. They often rely on rasterization or predefined spatial support to control the influence of Gaussian functions~\cite{gaussiansr,gsasr}. Such predefined support regions may limit the flexibility of pixel–Gaussian associations when modeling multi-scale objects and fine-grained spatial structures in remote sensing HSIs. In addition, Gaussian functions are often aggregated based on geometric proximity during pixel reconstruction. However, in remote sensing HSIs, spatially adjacent objects may have different spectral properties~\cite{su}. Therefore, aggregation based solely on spatial relationships does not explicitly account for feature consistency with the low-resolution input. Moreover, although Gaussian Splatting provides a continuous formulation for spatial reconstruction, preserving spectral fidelity remains critical in HSI super-resolution~\cite{srreview}.

To address these issues, we propose GaussianHSI for arbitrary-scale HSI super-resolution. We develop a Voronoi-Guided Bilateral 2D Gaussian Splatting (VBGS) module to improve the adaptive selection and fusion of Gaussian functions during pixel reconstruction, as illustrated in Fig.~\ref{fig:compare}. We also introduce a Spectral Detail Enhancement module for spectral recovery. The main contributions can be summarized as follows:
\begin{itemize}
    \item We propose GaussianHSI, a 2D Gaussian Splatting-based framework for arbitrary-scale HSI super-resolution. To the best of our knowledge, this is the first attempt to introduce Gaussian Splatting into arbitrary-scale HSI super-resolution.

    \item We develop a Voronoi-guided Gaussian Selection strategy that identifies the most relevant Gaussian functions for each target pixel based on anisotropic spatial relevance, making the selection more adaptive to diverse spatial structures.

    \item We introduce a Reference-aware Gaussian Aggregation mechanism that uses low-resolution features as a reference and considers both geometric relevance and feature consistency during reconstruction.

    \item Extensive experiments on benchmark datasets demonstrate the effectiveness of the proposed method for both single-scale and arbitrary-scale HSI super-resolution.
    
\end{itemize}

\section{Method}
\begin{figure*}
    \centering
    \includegraphics[width=0.99\linewidth]{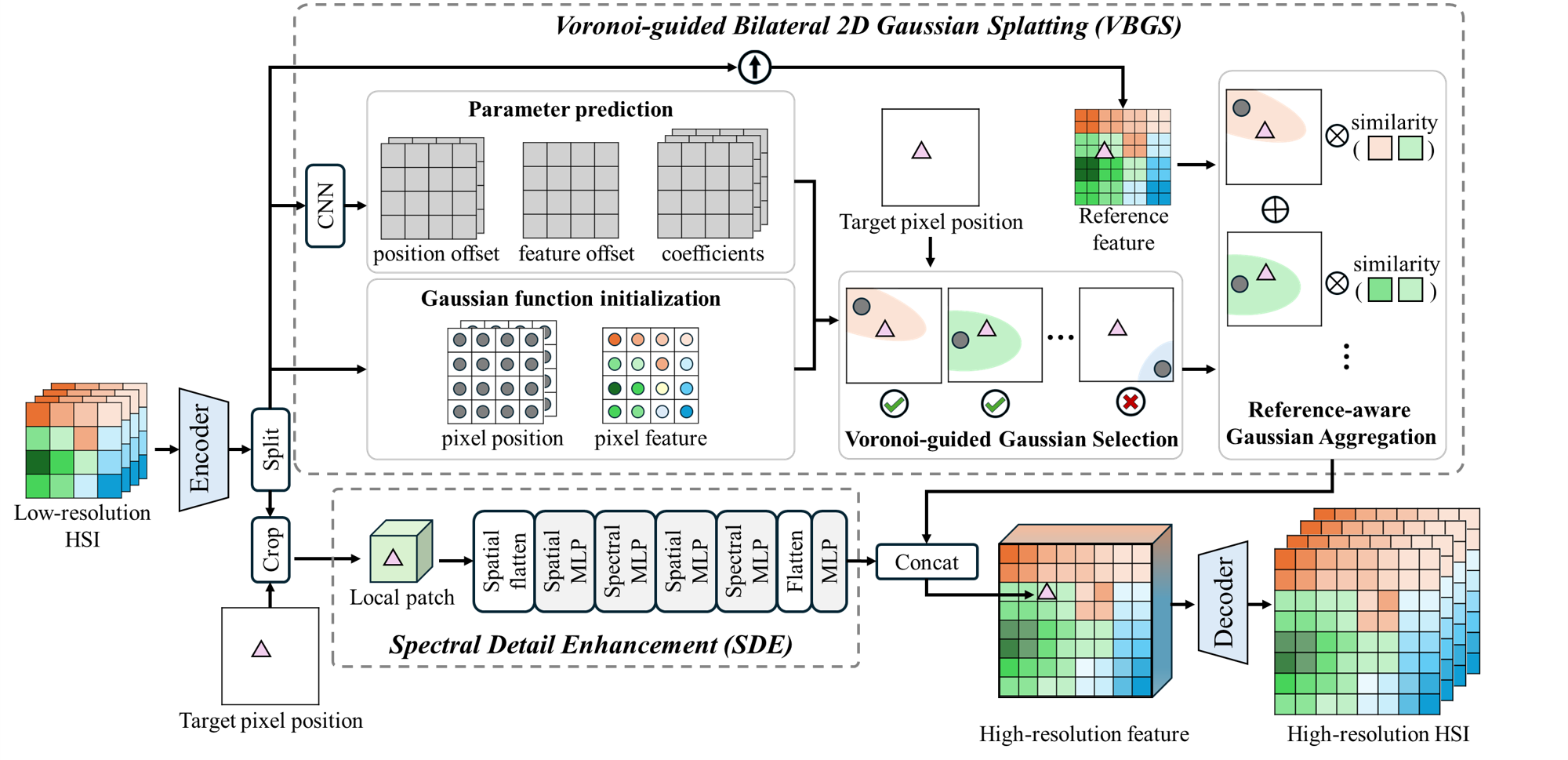}
    \caption{Overview of the proposed GaussianHSI. The low-resolution HSI is first encoded and split into a VBGS module and an SDE module. The VBGS performs spatial reconstruction, while the SDE enhances spectral detail recovery. The features from the two branches are then concatenated and decoded to reconstruct the high-resolution HSI.}
    \label{fig:model}
\end{figure*}

\subsection{Preliminaries: 2D Gaussian Splatting}
2D Gaussian splatting represents image content using a collection of spatially distributed Gaussian functions~\cite{gaussianimage,gsasr}. Each Gaussian function describes a spatial distribution with associated color coefficients. By aggregating the contributions of these Gaussian functions, pixel values can be computed at given spatial locations. This allows reconstruction on pixel grids at different resolutions, making the representation suitable for arbitrary-scale image super-resolution~\cite{gaussiansr,gsasr}. Specifically, an image is represented by a set of Gaussian functions parameterized as: 
\begin{align}
    \mathcal{G} = \{(x_{n}, y_n, c_n, o_n,  \Sigma_{n} )\}_{n=1}^{N},
\end{align}
\noindent 
where $x_n, y_n \in \mathbb{R}$ denote the horizontal and vertical coordinates of the $n$-th Gaussian center, respectively, $c_n \in \mathbb{R}^{C}$ denotes its associated $C$-channel color vector, $o_n \in \mathbb{R}$ denotes its opacity coefficient, $\Sigma_n \in \mathbb{R}^{2\times 2}$ is the covariance matrix, and $N$ denotes the number of Gaussian functions. The covariance matrix describes the spatial extent and orientation of the Gaussian function, which is parameterized by $\sigma_{x,n} > 0, \sigma_{y,n} > 0, \rho_n \in (-1, 1)$ as
\begin{equation}
    \Sigma_n = 
    \begin{bmatrix}
\sigma_{x,n}^2 & \rho_n \sigma_{x,n}\sigma_{y,n} \\
\rho_n \sigma_{x,n}\sigma_{y,n} & \sigma_{y,n}^2
\end{bmatrix},
\end{equation}
\noindent 
where $\sigma_{x,n}$ and $\sigma_{y,n}$ are the standard deviations along the horizontal and vertical axes, respectively, controlling the spatial extent along each axis. The coefficient $\rho_n$ describes the correlation between the two spatial components. Given the Gaussian representation $\mathcal{G}$ of the image, the contribution of the $n$-th Gaussian function at location $(x,y)$ is defined as
\begin{align}
    G_n(x,y)=\frac{c_n\cdot o_n}{2\pi |\Sigma_n|^{1/2}}
    \exp\left(
    -\frac{1}{2}
    \begin{bmatrix}
        x-x_n\\
        y-y_n
    \end{bmatrix}^{\text{T}}
    \Sigma_n^{-1}
    \begin{bmatrix}
        x-x_n\\
        y-y_n
    \end{bmatrix}
    \right).
    \label{gf}
\end{align}
\noindent
The image value at location $(x,y)$ is then obtained by aggregating the contributions of all Gaussian functions:
\begin{align}
    I(x,y)=\sum_{n=1}^{N} G_n(x,y),
\end{align}
\noindent
where $I(x,y) \in \mathbb{R}^C$ denotes the pixel value at location $(x,y)$. This formulation provides a continuous image representation. Images at different resolutions can therefore be reconstructed by evaluating the representation at the corresponding target pixel locations, making it suitable for arbitrary-scale super-resolution. Direct evaluation requires aggregating all Gaussian functions for each target pixel. This can incur substantial computational cost, including redundant computation for Gaussian functions with negligible contributions~\cite{continuoussr,gsasr}. To improve efficiency, existing methods often use rasterization strategies that compute each Gaussian function's contributions only within a fixed neighboring region~\cite{gaussiansr,gsasr,continuoussr}. However, rasterization strategies based on predefined spatial ranges may have limited flexibility in adapting pixel-Gaussian associations to Gaussian functions with different spatial extents and orientations. In addition, aggregation based solely on spatial relevance does not explicitly consider feature consistency between Gaussian features and low-resolution inputs.

\subsection{Overview}
Although 2D Gaussian splatting provides a flexible continuous representation and has shown promising performance in color image super-resolution, extending it to HSI super-resolution requires careful consideration of both spatial structure reconstruction and spectral fidelity. To this end, we propose GaussianHSI for arbitrary-scale HSI super-resolution. As illustrated in Fig.~\ref{fig:model}, GaussianHSI first extracts latent features from the input HSI. High-resolution spatial features are reconstructed using the proposed Voronoi-guided Bilateral 2D Gaussian Splatting (VBGS) module, while a parallel Spectral Detail Enhancement (SDE) module enhances spectral detail recovery. Finally, the outputs of the two modules are concatenated and decoded to generate the high-resolution HSI. The proposed GaussianHSI is motivated by three considerations. First, while existing rasterization strategies often use local neighborhoods to limit Gaussian aggregation for efficiency, we introduce Voronoi-guided selection to allow more flexible pixel-Gaussian associations~\cite{voronoi,maha}. By selecting Gaussian functions according to anisotropic distances that reflect their learned spatial extents and orientations, this strategy allows each target pixel to incorporate spatially relevant Gaussian functions beyond a predefined local neighborhood. Second, geometric relevance alone does not explicitly account for feature consistency with the low-resolution input. We therefore use upsampled low-resolution features as a reference to modulate the contributions of the selected Gaussian functions according to feature similarity~\cite{mm1}. The resulting aggregation jointly considers geometric relevance and feature consistency, providing additional guidance from the input for spatial reconstruction. Third, Gaussian-based reconstruction primarily focuses on spatial structure modeling, while HSI super-resolution also requires preserving spectral relationships. We therefore introduce SDE to complement VBGS in recovering fine spectral details.

Given a low-resolution HSI $\mathcal{I}^{lr} \in \mathbb{R}^{H\times W\times B}$, where $H\times W$ and $B$ denote the spatial size and number of spectral bands, respectively, the goal is to reconstruct a high-resolution HSI $\mathcal{I}^{hr} \in \mathbb{R}^{rH\times rW \times B}$ with a scale factor $r$. The target spatial dimensions are rounded to integers for non-integer scale factors. The overall mapping of GaussianHSI is formulated as:
\begin{equation}
    \mathcal{I}^{hr} = \text{GaussianHSI}(\mathcal{I}^{lr}, r),
\end{equation}
\noindent
where $\text{GaussianHSI}(\cdot)$ denotes the proposed method. We first employ an encoder to extract features from the input HSI. In the encoder, the input is first projected into a low-dimensional latent space through a pixel-wise fully connected layer, followed by four residual convolutional blocks~\cite{edsr}. Each block consists of two convolution layers with a rectified linear unit activation in between and a residual connection from the input to the output. The encoded feature is denoted as
\begin{align}
    \mathcal{F}_{en} = \text{En}(\mathcal{I}^{lr}),
\end{align}
\noindent
where $\mathrm{En}(\cdot)$ denotes the encoder and $\mathcal{F}_{en} \in \mathbb{R}^{H\times W\times C_{en}}$ is the extracted latent feature. We then split $\mathcal{F}_{en}$ along the channel dimension into two parts, namely the first $C_1$ channels and the remaining $C_2$ channels:
\begin{align}[\mathcal{F}_{spatial},\mathcal{F}_{spectral}] = \text{Split}(\mathcal{F}_{en}), 
\end{align}
\noindent
where $\mathcal{F}_{spatial} \in  \mathbb{R}^{H\times W\times C_1}$ is used for spatial feature reconstruction and $\mathcal{F}_{spectral} \in  \mathbb{R}^{H\times W\times C_2}$ is used for spectral detail enhancement, with $C_1 + C_2 = C_{en}$. For spatial feature reconstruction, we employ the proposed VBGS module:
\begin{align}
    \mathcal{F}'_{spatial} = \text{VBGS}(\mathcal{F}_{spatial}, r),
\end{align}
\noindent
where $\mathcal{F}'_{spatial} \in \mathbb{R}^{rH\times rW\times C_1}$ denotes the restored high-resolution feature, and $\text{VBGS}(\cdot)$ denotes the proposed VBGS module. In parallel, we adopt the SDE module to enhance spectral details:
\begin{align}
    \mathcal{F}'_{spectral} = \text{SDE}(\mathcal{F}_{spectral} \uparrow_{r}),
\end{align}
\noindent
where $\mathcal{F}'_{\text{spectral}} \in \mathbb{R}^{rH \times rW \times C_2}$ denotes the enhanced spectral feature, $\uparrow_{r}$ denotes spatial upsampling with scale factor $r$, and $\mathrm{SDE}(\cdot)$ denotes the proposed module. Detailed descriptions of VBGS and SDE are provided in the following subsections. The outputs of the two modules are concatenated along the channel dimension and then fed into a decoder for HSI reconstruction:
\begin{align}
    \mathcal{F}' &= \mathrm{Concat}(\mathcal{F}'_{spatial}, \mathcal{F}'_{spectral}), \\
    \mathcal{I}^{hr} &= \text{De}(\mathcal{F}'),
\end{align}
\noindent
where $\mathcal{F}' \in \mathbb{R}^{rH \times rW \times C_{en}}$ denotes the fused high-resolution latent feature, and $\mathrm{De}(\cdot)$ denotes the decoder. The decoder partitions the fused feature map into local windows and converts them into a sequence of tokens. Self-attention is applied across these tokens to integrate spatial information~\cite{vit}. The resulting tokens are then rearranged into a feature map with a spatial size $rH\times rW$, and a pixel-wise fully connected layer maps each pixel feature to its corresponding spectral vector to reconstruct the high-resolution HSI.

\begin{algorithm}[t]
\caption{VBGS Feature Reconstruction}
\label{alg:vbgs}
\begin{algorithmic}[1]
\REQUIRE Low-resolution spatial feature $\mathcal{F}_{spatial}$, scale factor $r$, top-$k$
\ENSURE Reconstructed high-resolution feature $\mathcal{F}'_{spatial}$

\STATE \textit{// Gaussian parameter prediction.}
\STATE Predict Gaussian parameters using Eqs.~(\textcolor{blue}{12})-(\textcolor{blue}{17}).

\item[]

\STATE \textit{// Gaussian primitive construction.}
\STATE Construct Gaussian functions $G_{i,j}(x,y)$ from the predicted parameters.

\item[]

\STATE \textit{// All high-resolution pixels are reconstructed in parallel.}
\STATE \textit{// Parallel Voronoi-guided Gaussian selection.}
\STATE Compute Mahalanobis distances $d_{i,j}(x,y) $. \hfill \hfill // Eq.~(\textcolor{blue}{20})
\STATE Select $k$ Gaussian functions for each pixel to form $\Omega_k(x,y)$. \\ \hfill // Eq.~{\color{blue}(21)-(22)}

\item[]

\STATE \textit{// Parallel reference-aware bilateral aggregation.}
\STATE Upsample $\mathcal{F}_{spatial}$ to obtain the reference features.
\STATE Compute weights $w_{i,j}(x,y)$ for $(i,j) \in \Omega_k(x,y)$. \\ \hfill  // Eq.~{\color{blue}(23)-(24)}
\STATE Compute Gaussian contributions $G_{i,j}(x,y)$ for $(i,j) \in \Omega_k(x,y)$. \\ \hfill // Eq.~(\textcolor{blue}{18})

\STATE Reconstruct $\mathcal{F}'_{spatial}(x,y)$ by normalized weighted aggregation over $(i,j)\in\Omega_k(x,y)$. \\ \hfill // Eq.~(\textcolor{blue}{25})

\RETURN $\mathcal{F}'_{spatial}$
\end{algorithmic}
\end{algorithm}

\subsection{Voronoi-Guided Bilateral 2D Gaussian Splatting}
We develop a Voronoi-guided Bilateral 2D Gaussian Splatting (VBGS) module for spatial reconstruction. Given the input feature map, VBGS first predicts the parameters of a set of Gaussian functions to form a continuous spatial representation. For each target high-resolution pixel, Voronoi-guided selection identifies the top-$k$ Gaussian functions with the smallest anisotropic distances. The selected Gaussian functions are then weighted using the upsampled low-resolution features as a reference, with the weights determined by the similarity between the Gaussian features and the reference features. This reference-aware bilateral aggregation combines geometric relevance with feature consistency to reconstruct the target pixel feature. The overall pipeline of VBGS is summarized in Algorithm~\ref{alg:vbgs}.

\noindent \textbf{Initialization and Parameterization.} Given the low-resolution feature map $\mathcal{F}_{spatial} \in \mathbb{R}^{H \times W \times C_1}$, we initialize one 2D Gaussian function at each low-resolution pixel location. Remote sensing HSIs often contain small objects and fine spatial structures~\cite{tobler}. Assigning a Gaussian function to each location provides dense spatial coverage of the input feature map as a starting point for modeling these details. Compared with directly regressing Gaussian parameters from scratch, this initialization provides a structured parameterization strategy that is beneficial for stable optimization~\cite{continuoussr,gaussiansr}. We then use a set of independent three-layer CNNs, denoted by $\{\Phi_{\xi}(\cdot)\}_{\xi \in \{X,Y,F,\sigma_x,\sigma_y,\rho\}}$, to predict the parameters defining the Gaussian centers, associated coefficients, and covariance matrices. The Gaussian center positions are predicted as
\begin{align}
    X &= X_{init} + \mathrm{Tanh}(\Phi_{X}(\mathcal{F}_{spatial})), \\
    Y &= Y_{init} + \mathrm{Tanh}(\Phi_{Y}(\mathcal{F}_{spatial})),
\end{align}
\noindent
where $X, Y \in \mathbb{R}^{H \times W \times 1}$ denote the final horizontal and vertical coordinates of the Gaussian centers, respectively. $X_{init}, Y_{init} \in \mathbb{R}^{H \times W \times 1}$ denote the corresponding initial coordinates normalized to $[-1,1]$, and $\mathrm{Tanh}(\cdot)$ denotes the hyperbolic tangent activation function. The feature associated with each Gaussian function is defined as
\begin{align}
    F &= \mathrm{ReLU}(\mathcal{F}_{spatial} + \Phi_{F}(\mathcal{F}_{spatial})),
\end{align}
where $F \in \mathbb{R}^{H \times W \times C_1}$ denotes the Gaussian feature, and $\mathrm{ReLU}(\cdot)$ denotes the rectified linear unit activation function. For simplicity, the color and opacity terms are combined into a single Gaussian feature vector $F_{i,j}$ for each Gaussian function. The covariance coefficients are predicted as:
\begin{align}
    \sigma_x &= \text{Softplus}(\Phi_{\sigma_x}(\mathcal{F}_{spatial})), \\
    \sigma_y &= \text{Softplus}(\Phi_{\sigma_y}(\mathcal{F}_{spatial})), \\
    \rho   &= \mathrm{Tanh}(\Phi_{\rho}(\mathcal{F}_{spatial})),
\end{align}
where $\sigma_x, \sigma_y \in \mathbb{R}^{H \times W \times 1}$ denote the standard deviations along the horizontal and vertical axes, respectively, and $\rho \in \mathbb{R}^{H \times W \times 1}$ denotes the correlation coefficient. $\mathrm{Softplus}(\cdot)$ denotes the softplus activation function, which ensures positive standard deviations. Using these predicted parameters, the contribution of the Gaussian function indexed by $(i, j)$ to the target pixel location $(x,y)$ is formulated, following Eq.~\ref{gf}, as
\begin{align}
    G_{i,j}&(x,y) = \\ \nonumber
    & \frac{F_{i,j}}{2\pi |\Sigma_{i,j}|^{1/2}}
    \exp\left(
    -\frac{1}{2}
    \begin{bmatrix}
        x-X_{i,j}\\
        y-Y_{i,j}
    \end{bmatrix}^{\text{T}}
    \Sigma_{i,j}^{-1}
    \begin{bmatrix}
        x-X_{i,j}\\
        y-Y_{i,j}
    \end{bmatrix}
    \right),
\end{align}
\noindent
where
\begin{align}
        \Sigma_{i,j} = 
    \begin{bmatrix}
\sigma_{x,ij}^2 & \rho_{i,j} \sigma_{x,ij}\sigma_{y,ij} \\
\rho_{i,j} \sigma_{x,ij}\sigma_{y,ij} & \sigma_{y,ij}^2
\end{bmatrix},
\end{align}
\noindent
and the target pixel coordinates $(x,y)$ are also normalized to $[-1,1]$

\noindent \textbf{Voronoi-Guided Gaussian Selection.}
After obtaining the Gaussian functions, we reconstruct the feature at each target high-resolution pixel location by aggregating contributions from selected Gaussian functions. Direct aggregation over all Gaussian functions incurs substantial computational cost, including redundant computation for Gaussian functions with negligible contributions to the target pixel~\cite{gsasr}. To improve efficiency, existing methods often adopt rasterization strategies that restrict the Gaussian functions considered for aggregation to a predefined local region~\cite{continuoussr}. However, predefined local regions may offer limited adaptability in establishing pixel-Gaussian associations under varying Gaussian spatial extents and orientations. To accommodate these spatial characteristics, VBGS introduces Voronoi-guided Gaussian selection. Inspired by the proximity-based assignment principle of Voronoi diagrams~\cite{voronoi}, this strategy evaluates pixel-Gaussian relevance using Mahalanobis distances. These distances account for the relative position of each target pixel with respect to a Gaussian center, as well as the Gaussian's spatial extent and orientation. For each target pixel, the top-$k$ Gaussian functions with the smallest distances are retained for aggregation. The selected set thus adapts to the spatial relevance of the Gaussian functions without being restricted to a predefined local neighborhood. Specifically, for a target pixel location $(x,y)$ and the Gaussian function indexed by $(i,j)$, the relevance is measured using the Mahalanobis distance:
\begin{align}
    d_{i,j}(x,y)=
    \sqrt{
    \begin{bmatrix}
        x-X_{i,j}\\
        y-Y_{i,j}
    \end{bmatrix}^{\top}
    \Sigma_{i,j}^{-1}
    \begin{bmatrix}
        x-X_{i,j}\\
        y-Y_{i,j}
    \end{bmatrix}
    }.
\end{align}
Compared with Euclidean distance, the Mahalanobis distance better reflects the scale and orientation of each Gaussian function. Based on this relevance, we select the top-$k$ Gaussian functions with the smallest distances to form the candidate Gaussian set:
\begin{align}
    \Omega_{k}(x,y)
    &= \operatorname*{arg\,mink}_{(i,j)} d_{i,j}(x,y), \\
    \mathcal{G}_{k}(x,y)
    &= \left\{ G_{i,j} \mid (i,j) \in \Omega_{k}(x,y) \right\},
\end{align}
\noindent
where $\text{arg mink}$ returns the indices corresponding to the $k$ smallest values, $\Omega_{k}(x,y)$ denotes the selected index set for target pixel $(x,y)$, and $\mathcal{G}_{k}(x,y)$ denotes the resulting Gaussian set for target pixel $(x,y)$. In this way, VBGS retains the $k$ most spatially relevant Gaussian functions under the Mahalanobis distance measure for each target pixel, enabling subsequent aggregation without predefined local support.

\noindent \textbf{Reference-Aware Bilateral Aggregation.}
After obtaining the candidate Gaussian set $\mathcal{G}_{k}(x,y)$ for each target pixel, we incorporate low-resolution features as a reference for aggregation. Geometric relevance characterizes the spatial relationship between a Gaussian function and the target pixel. However, spatially relevant Gaussian functions may represent different local content, particularly near boundaries between regions with distinct spectral properties. Feature consistency with the low-resolution input can therefore provide complementary guidance for aggregation. VBGS uses the upsampled low-resolution feature at the target location as the reference feature. The similarity between each selected Gaussian feature and the reference feature determines its aggregation weight. The Gaussian contributions are then combined through a normalized weighted sum to reconstruct the target pixel feature. This reference-aware bilateral aggregation jointly considers geometric relevance and feature consistency during spatial reconstruction. Specifically, the low-resolution spatial feature map $\mathcal{F}_{spatial}$ is upsampled to the target resolution. The reference feature at location $(x,y)$ is denoted by $\hat{\mathcal{F}}_{spatial}^{x,y} \in\mathbb{R}^{C_1}$. For each selected Gaussian function in $\mathcal{G}_{k}(x,y)$, we compute the cosine similarity between its feature vector and the reference feature:
\begin{align}
    s_{i,j}(x,y)
    =
    \frac{
    \left\langle F_{i,j}, \hat{\mathcal{F}}_{spatial}^{x,y} \right\rangle
    }{
    \|F_{i,j}\|_2 \, \|\hat{\mathcal{F}}_{spatial}^{x,y}\|_2 +\epsilon
    },
\end{align}
where $\epsilon>0$ is a small constant added to avoid division by zero. We further define a reference-aware weight as
\begin{align}
    w_{i,j}(x,y)=\exp(\gamma s_{i,j}(x,y)),
\end{align}
where $\gamma$ is a learnable coefficient controlling the influence of the reference feature. The reconstructed spatial feature at the target pixel $(x,y)$ is given by
\begin{align}
    \mathcal{F}'_{spatial}(x,y)
    =
    \frac{
    \sum_{G_{i,j}\in\mathcal{G}_{k}(x,y)}
    w_{i,j}(x,y)\, G_{i,j}(x,y)
    }{
    \sum_{G_{i,j}\in\mathcal{G}_{k}(x,y)}
    w_{i,j}(x,y)
    },
\end{align}
\noindent
where $\mathcal{F}'_{spatial}(x,y)\in\mathbb{R}^{C_1}$ denotes the reconstructed feature at the location $(x,y)$.

\begin{figure*}[!ht]
    \centering
    \includegraphics[width=1\linewidth]{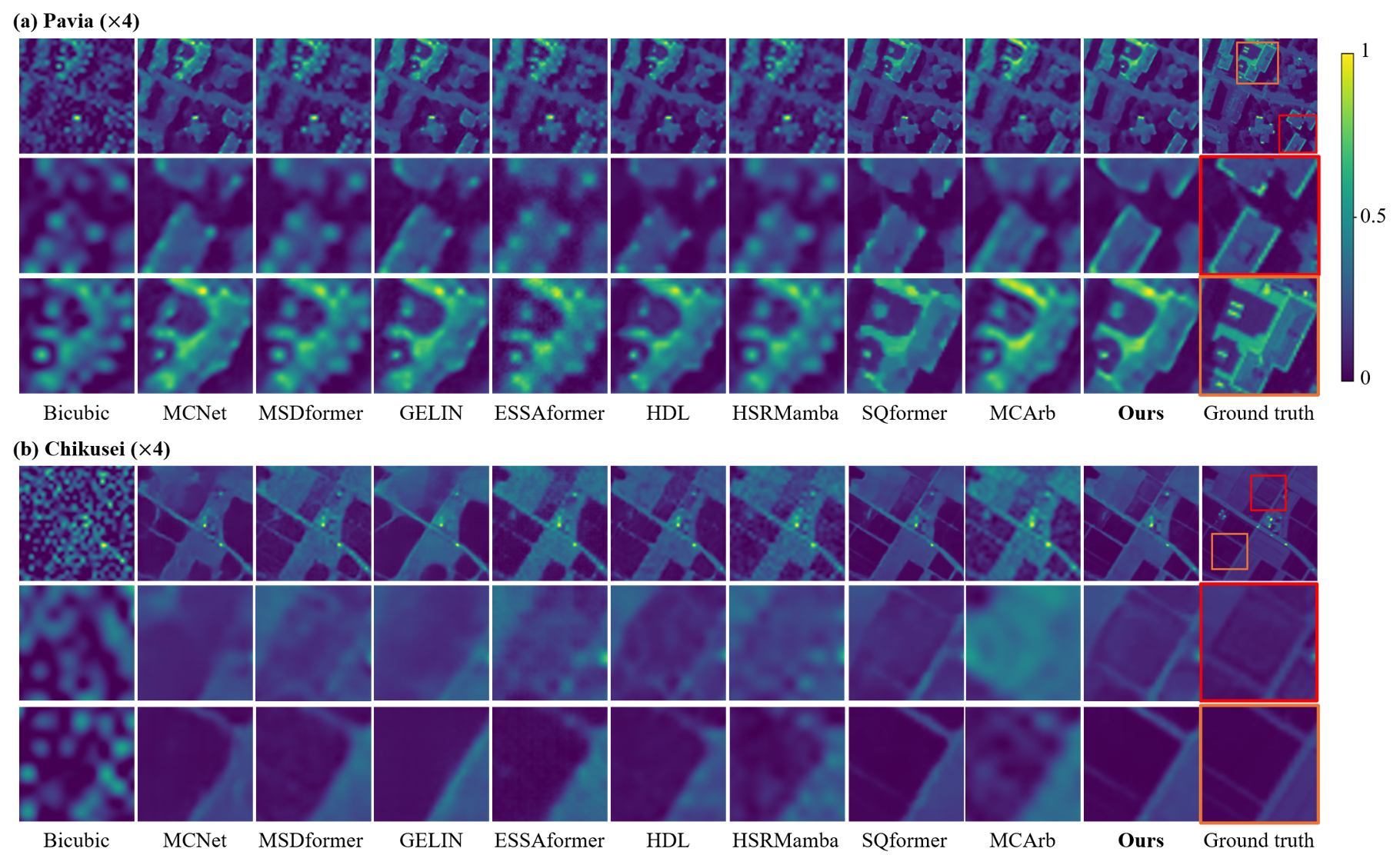}
    \caption{Qualitative comparison of single-scale HSI super-resolution results on the Pavia and Chikusei datasets at scale $\times 4$. The 23rd band is visualized in pseudo color. The first and fourth rows show the restored results. The second and fifth rows provide zoomed-in comparisons of the regions marked by red boxes, while the third and sixth rows provide zoomed-in comparisons of the regions marked by orange boxes.}
    \label{fig:fixres}
\end{figure*}

\subsection{Spectral Detail Enhancement}
In parallel with the VBGS, we introduce a Spectral Detail Enhancement (SDE) module to enhance spectral reconstruction. Given the low-resolution spectral feature $\mathcal{F}_{spectral}\in\mathbb{R}^{H\times W\times C_2}$, we first upsample it to the target resolution:
\begin{align}
    \widetilde{\mathcal{F}}_{spectral}
    =
    \mathcal{F}_{spectral}\uparrow_{r},
\end{align}
\noindent
where $\widetilde{\mathcal{F}}_{spectral}\in\mathbb{R}^{rH\times rW\times C_2}$ denotes the upsampled spectral feature. For each target pixel location $(x,y)$, we extract a local patch $L_{x,y}\in\mathbb{R}^{P\times P\times C_2}$ centered at $(x,y)$ from $\widetilde{\mathcal{F}}_{spectral}$. Reflection padding is applied to extract local patches centered at pixels near feature map boundaries. The extracted patch is then processed by two layers of alternating spatial and spectral transformations:
\begin{align}
    L^{1}_{x,y} &= \Psi^{1}_{spe}(\Psi^{1}_{spa}(L_{x,y})), \\
    L^{2}_{x,y} &= \Psi^{2}_{spe}(\Psi^{2}_{spa}(L^{1}_{x,y})),
\end{align}
\noindent
where $\Psi^{l}_{spa}(\cdot)$ and $\Psi^{l}_{spe}(\cdot)$ denote the spatial multilayer perceptron (MLP)~\cite{hypermlp} and spectral MLP at the $l$-th layer, respectively. Specifically, each extracted local patch is reshaped into a set of spatial tokens, where each token corresponds to one spatial location with its spectral feature vector. The spatial MLP performs token mixing across spatial locations, and the spectral MLP performs channel mixing on the spectral feature vector of each token. Finally, the refined local patch feature at $(x,y)$ is flattened into a vector and fused to obtain the enhanced spectral feature:
\begin{align}
    \mathcal{F}'_{spectral}(x,y)
    =
    \Psi_{fuse}(\text{flatten}(L^2_{x,y})),
\end{align}
\noindent
where $\mathcal{F}'_{spectral}(x,y)\in\mathbb{R}^{C_2}$ denotes the enhanced spectral feature, $\Psi_{fuse}(\cdot)$ denotes the fusion MLP, and $\text{flatten}(\cdot)$ denotes vectorization.

\section{Experiments}
\subsection{Experimental Setup}
\noindent \textbf{Dataset.} We evaluate the proposed method on three widely used remote-sensing HSI datasets: Houston~\footnote{https://www.grss-ieee.org/community/technical-committees/2013-ieee-grss-data-fusion-contest/}, Pavia \cite{pavia}, and Chikusei \cite{chikusei}. The Houston dataset was acquired by the ITRES-CASI 1500 sensor over Houston in 2012. It has a spatial size of $349 \times 1905$ pixels and contains $144$ spectral bands. The Pavia dataset was collected by the ROSIS sensor over Pavia in northern Italy. It comprises $1096 \times 715$ pixels with $102$ spectral bands. The Chikusei dataset was captured by the Hyperspec-VNIR-C sensor over Chikusei, Ibaraki, Japan. It has a spatial size of $2517 \times 2335$ pixels and contains $128$ spectral bands. For each dataset, we divide the full-resolution HSI into non-overlapping patches of $120 \times 120$ pixels and use half of these patches for training. Following \cite{msdformer,sqformer}, we generate the corresponding low-resolution data by downsampling the high-resolution data.

\noindent \textbf{Implementation Details.} All convolutional layers in the encoder use a kernel size of 3, with 128 hidden channels and 64 output channels. The input feature dimensions of VBGS and SDE are set to 16 and 48, respectively. In VBGS, the parameter prediction network uses convolutional layers with a kernel size of 3 and 64 hidden channels. The top-16 relevant Gaussian functions are selected for pixel reconstruction. In SDE, $P$ is set to 5. Both the spatial and spectral MLPs consist of two fully connected layers with ReLU activations, and the hidden dimension is set to 64. High-resolution pixels are calculated in parallel for efficient training and inference. In the decoder, the local window size is set to 24, and 4-head attention is used. The model is trained for 500 epochs using the Adam optimizer~\cite{adam} with an $\ell_1$ reconstruction loss. The batch size is set to 1. The learning rate is initialized to 0.0008 and reduced to 0.0001 after 100 epochs. Our model is implemented in PyTorch and trained on an NVIDIA RTX A6000 GPU. 

\begin{table*}[!ht]
    \centering
    \caption{Quantitative comparison of single-scale HSI super-resolution results with state-of-the-art methods.}
    \begin{tabular}{c | c | c | c c c c c c c c c | c}
    \toprule
    Dataset & Scale & Metrics & Bicubic & MCNet & MSDformer & GELIN & ESSAformer & HDL & HSRMamba & SQformer & MCArb  & Ours \\ \hline
    \multirow{9}*{Houston}& \multirow{3}*{$\times 2$}  & PSNR$\uparrow$ & 29.53 & 36.74 & 36.73 & - & 35.89 & 36.41 & 36.91 & \underline{37.07}&36.39 & \textbf{37.12}  \\
    &  & SSIM$\uparrow$ & 0.599 & 0.917 & 0.919 & - & 0.904 & 0.917 & 0.920 & \underline{0.922}& 0.913  & \textbf{0.929}  \\
    &  & SAM$\downarrow$ & 0.385 & 0.088 & 0.081 & - & 0.098 & 0.086 & 0.078 & \underline{0.074} & 0.081 & \textbf{0.069}  \\ \cline{2-13}
    & \multirow{3}*{$\times 4$} & PSNR$\uparrow$ & 28.11 & 33.20 & 33.17 & 33.36 & 32.44 & 33.06 & 33.46 & \underline{33.54}  & 33.19 & \textbf{33.75}  \\
    &  & SSIM$\uparrow$ & 0.572 & 0.822 & 0.821 & 0.826 & 0.799 & 0.820 & 0.826 & \underline{0.829} &  0.822 & \textbf{0.833}  \\
    &  & SAM$\downarrow$ & 0.396 & 0.134 & 0.119 & 0.128 &0.131  & 0.117 & 0.114 & \underline{0.106}& 0.114 & \textbf{0.101}  \\ \cline{2-13}
    & \multirow{3}*{$\times 8$} & PSNR$\uparrow$ & 26.89 & 30.59 & 30.68 & 31.09 & 30.32 & 30.51 & 30.64 & \underline{31.15}& 30.37 & \textbf{31.30}  \\
    &  & SSIM$\uparrow$ & 0.559 & 0.728 & 0.729 & 0.741 & 0.722 & 0.738 & 0.731 & \underline{0.752}& 0.728  & \textbf{0.760}  \\ 
    &  & SAM$\downarrow$ & 0.418 & 0.193 & 0.165 & 0.171 & 0.169 & 0.166 & 0.160 & \underline{0.156}& 0.171  & \textbf{0.152}  \\\hline

    \multirow{9}*{Pavia}& \multirow{3}*{$\times 2$} & PSNR$\uparrow$ & 28.21 & 33.63 & 33.26 & - & 32.05 & 32.71 & 33.31 & \underline{33.91} & 33.08  & \textbf{34.20}   \\
    &  & SSIM$\uparrow$ & 0.722 & 0.925 & 0.921 & - & 0.903 & 0.909 & 0.923 & \underline{0.928} & 0.919 & \textbf{0.933}   \\
    &  & SAM$\downarrow$ & 0.293 & 0.097 & 0.099 & - & 0.106 & 0.114 & 0.100 & 0.096 & \underline{0.095} & \textbf{0.090}   \\ \cline{2-13}
    & \multirow{3}*{$\times 4$} & PSNR$\uparrow$ & 25.28 & 27.85 & 27.61 & 28.11 & 26.96 & 27.53 & 27.56 & \underline{28.26}&  27.75  & \textbf{28.57}   \\
    &  & SSIM$\uparrow$ & 0.591 & 0.744 & 0.738 & 0.757 & 0.717 & 0.737 & 0.734 & \underline{0.761} & 0.746 & \textbf{0.769}   \\
    &  & SAM$\downarrow$ & 0.308 & 0.147 & 0.135 & 0.140 & 0.139 & 0.142 & 0.135 & \underline{0.129}& 0.134 & \textbf{0.124}   \\ \cline{2-13}
    & \multirow{3}*{$\times 8$} & PSNR$\uparrow$ & 22.62 & 24.63 & 24.59 & \underline{25.24} & 24.03 & 23.94 & 24.08 & 25.20& 24.53 & \textbf{25.47}   \\
    &  & SSIM$\uparrow$ & 0.438 & 0.534 & 0.531 & \underline{0.575} & 0.508 & 0.501 & 0.511 & 0.572 & 0.547 & \textbf{0.580}   \\
    &  & SAM$\downarrow$ & 0.327 & 0.186 & 0.177 & 0.183 & 0.177 & 0.197 & 0.172 & \underline{0.167} & 0.171 & \textbf{0.164}  \\ \hline

    \multirow{9}*{Chikusei}& \multirow{3}*{$\times 2$} & PSNR$\uparrow$ & 30.26 & 42.23 & \underline{42.25} & - & 40.86 & 41.56 & 42.13 & 42.16 &  42.04& \textbf{42.48}   \\
    &  & SSIM$\uparrow$ & 0.598 & 0.968 & \underline{0.970} & - & 0.960 & 0.965 & 0.968 & 0.969&  0.961 & \textbf{0.973}   \\
    &  & SAM$\downarrow$ & 0.306 & 0.049 & \underline{0.047} & - & 0.056 & 0.051 & 0.050 & 0.048 & 0.050 & \textbf{0.044}  \\ \cline{2-13}
    & \multirow{3}*{$\times 4$} & PSNR$\uparrow$ & 29.33 & 36.33 & \underline{36.53} & 36.07 & 35.47 & 35.86 & 35.53  & 36.48 & 36.17 &\textbf{36.70}  \\
    &  & SSIM$\uparrow$ & 0.592 & 0.896 & 0.901 & 0.887 & 0.874 & 0.892 & 0.878 & \underline{0.904}& 0.886 & \textbf{0.907}   \\
    &  & SAM$\downarrow$ & 0.312 & 0.072 & \underline{0.063} & 0.082 & 0.079 & 0.076 & 0.085  & 0.065 & 0.080 & \textbf{0.062}   \\ \cline{2-13}
    & \multirow{3}*{$\times 8$} & PSNR$\uparrow$ & 28.20 & 32.50 & \underline{32.56} & 32.51 & 32.10 & 32.18 & 32.44 & 32.42 & 32.29 &\textbf{32.61}   \\
    &  & SSIM$\uparrow$ & 0.582 & 0.810 & \underline{0.811} & 0.805 & 0.795 & 0.799 & 0.803 & 0.801 & 0.797 & \textbf{0.814}   \\
    &  & SAM$\downarrow$ & 0.320 & 0.106 & \textbf{0.100} & 0.111 & 0.115 & 0.110 & \underline{0.105} & 0.109 & 0.110 & \textbf{0.100}   \\
    \bottomrule
  \end{tabular}
    \label{tab:fixscale}
\end{table*}

\noindent \textbf{Evaluation metrics.} We evaluate performance using three widely used metrics: peak signal-to-noise ratio (PSNR), structural similarity index measure (SSIM), and spectral angle mapper (SAM). PSNR measures reconstruction fidelity based on pixel-wise errors, SSIM assesses the preservation of local spatial structures, and SAM evaluates spectral distortion through the angle between reconstructed and reference spectral vectors. Higher PSNR and SSIM values and lower SAM values indicate better performance.

\begin{figure*}[!t]
    \centering
    \includegraphics[width=0.999\linewidth]{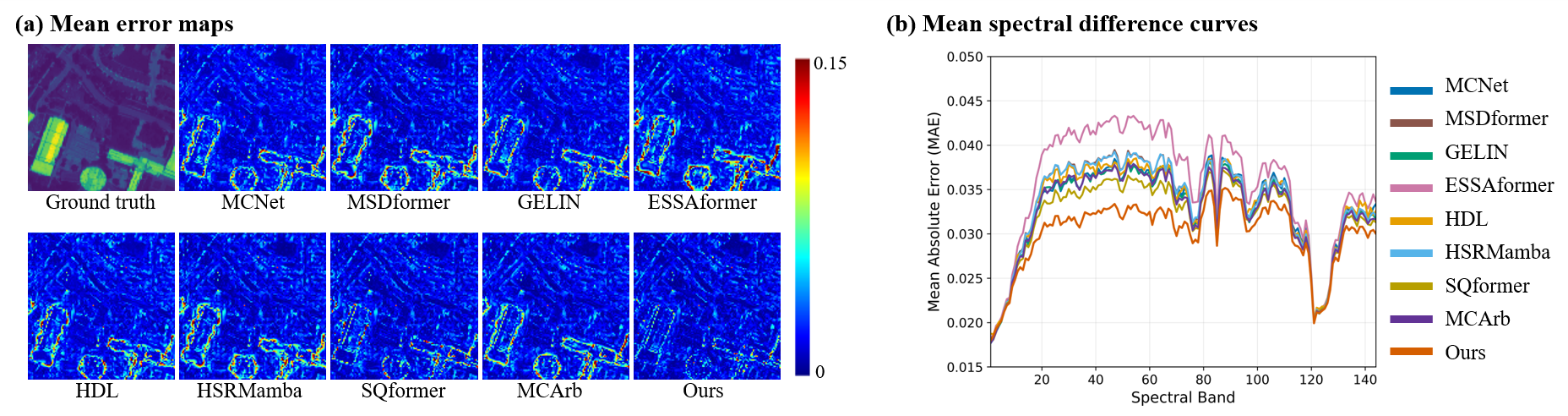}
    \caption{Comparison of reconstruction errors for different methods in single-scale HSI super-resolution on the Houston dataset at scale $\times$4. (a) Ground-truth image and spatial error maps obtained by averaging absolute errors over spectral bands. (b) Band-wise mean absolute error curves averaged over spatial pixels.}
    \label{fig:error_fix}
\end{figure*}

\subsection{Single-scale HSI super-resolution}
For single-scale evaluation, we compare the proposed method with representative HSI super-resolution methods, including MCNet~\cite{mcnet}, MSDformer~\cite{msdformer}, GELIN~\cite{gelin}, ESSAformer~\cite{essa}, HDL~\cite{hdl}, HSRMamba~\cite{hsrmamba}, SQformer~\cite{sqformer}, and MCArb~\cite{mcarb}. Following~\cite{hirdiff}, we add Gaussian noise with a noise level of 10 to the low-resolution HSIs to evaluate reconstruction performance under challenging noisy conditions. All methods are carefully tuned for the experimental setting and evaluated using the same data and evaluation protocol to ensure a fair comparison.

Quantitative results are reported in TABLE~\ref{tab:fixscale}. GaussianHSI achieves the highest PSNR and SSIM values across three datasets and evaluated scale factors, together with low SAM values. These results indicate its effectiveness in improving spatial reconstruction quality while preserving spectral fidelity. On the Pavia dataset, GaussianHSI improves PSNR over the best competing results by 0.29 dB, 0.31 dB, and 0.23~dB at scales $\times2$, $\times4$, and $\times8$, respectively. In addition, on the Houston dataset at a scale factor of $\times2$, GaussianHSI reduces SAM from 0.074 to 0.069 compared with SQformer, demonstrating improved spectral fidelity alongside the gain in PSNR. Similar improvements can be observed on the Chikusei dataset, where GaussianHSI achieves higher PSNR and SSIM while maintaining low spectral distortion. These results demonstrate the effectiveness of GaussianHSI for HSIs with different spatial and spectral characteristics.

\begin{figure*}[!t]
    \centering
    \includegraphics[width=0.985\linewidth]{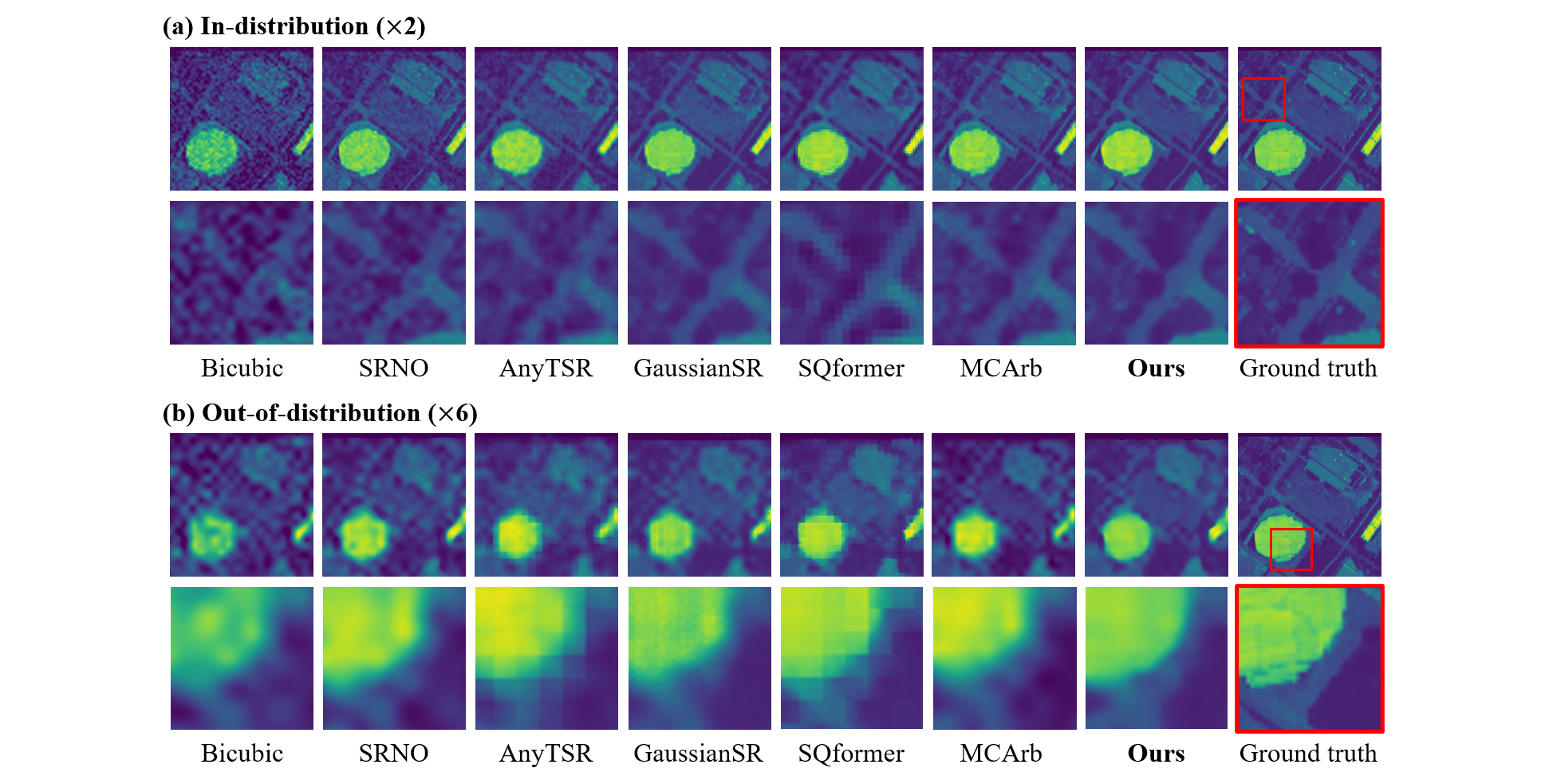}
    \caption{Qualitative comparison of arbitrary-scale HSI super-resolution results on the Houston dataset. The 23rd band is visualized in pseudo color. The first and third rows show the restored results. The second and fourth rows provide zoomed-in comparisons of the regions marked by the red box.}
    \label{fig:arbres}
\end{figure*}

\begin{table}[!t]
    \centering
    \caption{Quantitative results of arbitrary-scale HSI super-resolution on the Pavia dataset, including in-distribution scales ($\times 2$, $\times 4$) and out-of-distribution scales ($\times 6$, $\times 8$).}
    \begin{tabular}{@{\,}c@{\,}|c c c| c c c@{\,}}
    \toprule
        \rowcolor{gray!20} \multicolumn{7}{c}{\textbf{In-distribution}} \\
        \multirow{2}*{Method} & \multicolumn{3}{c}{$\times 2$} &  \multicolumn{3}{c}{$\times 4$}  \\ \cline{2-7}
        & PSNR$\uparrow$ & SSIM$\uparrow$ & SAM$\downarrow$ & PSNR$\uparrow$ & SSIM$\uparrow$ & SAM$\downarrow$ \\ \hline
        Bicubic & 28.21 & 0.722 & 0.293 & 25.28 & 0.591 & 0.308 \\
        SRNO  & 31.50 & 0.867 & 0.172 & 26.93 & 0.708 & 0.188 \\
        AnyTSR  & 31.76 & 0.880 & 0.154 & \underline{27.10} & 0.708 & 0.182 \\
        GaussianSR  & 32.06 & 0.890 & 0.124 & 26.55 & 0.688  & 0.150 \\
        SQformer  & \underline{32.12} & \underline{0.896} & \underline{0.115} & 26.57 & \underline{0.709} & \underline{0.146} \\
        MCArb  & 31.47 & 0.892 & 0.127 & 26.56 & 0.683 & 0.157 \\
        Ours  & \textbf{33.51} & \textbf{0.917} & \textbf{0.092} & \textbf{27.37} & \textbf{0.728} & \textbf{0.128} \\ \hline

        \rowcolor{gray!20} \multicolumn{7}{c}{\textbf{Out-of-distribution}} \\
        \multirow{2}*{Method} & \multicolumn{3}{c}{$\times 6$} &  \multicolumn{3}{c}{$\times 8$}  \\ \cline{2-7}
        & PSNR$\uparrow$ & SSIM$\uparrow$ & SAM$\downarrow$ & PSNR$\uparrow$ & SSIM$\uparrow$ & SAM$\downarrow$ \\ \hline
        Bicubic & 23.66 & 0.501 & 0.319 & 22.62 & 0.438 & 0.327 \\
        SRNO  &25.20  & 0.575 & 0.209 & \underline{24.57} & \underline{0.498} & 0.221 \\
        AnyTSR  & \underline{25.48} & 0.579 & 0.204 & 24.52 & 0.494 & 0.217 \\
        GaussianSR  & 24.66 & 0.551 & 0.201 & 24.08 & 0.464 & 0.213 \\
        SQformer  & 25.17 & \underline{0.584} & \underline{0.182} & 24.23 & \underline{0.498} & \underline{0.201} \\
        MCArb  & 24.62 & 0.557 & 0.183 & 23.54 & 0.490 & 0.208 \\
        Ours  & \textbf{25.68} & \textbf{0.599} & \textbf{0.157} & \textbf{24.74} & \textbf{0.506} & \textbf{0.174} \\ 
        \bottomrule
    \end{tabular}
    \label{arbres_pc}
\end{table}

Fig.~\ref{fig:fixres} presents qualitative comparisons on the Pavia and Chikusei datasets at scale $\times4$. Compared with existing methods, GaussianHSI reconstructs clearer edges and local structures, producing results that are visually closer to the ground truth in the highlighted regions. These observations are consistent with the improvements in PSNR and SSIM. Fig.~\ref{fig:error_fix} further compares reconstruction errors on the Houston dataset at scale $\times4$ through spatial error maps and band-wise mean absolute error curves. These comparisons evaluate the spatial distribution of reconstruction errors and their variation across spectral bands. GaussianHSI exhibits fewer regions with large reconstruction errors and lower mean absolute errors across most spectral bands. Overall, the quantitative and qualitative results demonstrate the effectiveness of GaussianHSI for single-scale HSI super-resolution.

\begin{table}[!t]
    \centering
    \caption{Quantitative results of arbitrary-scale HSI super-resolution on the Houston dataset, including in-distribution scales ($\times 2$, $\times 4$) and out-of-distribution scales ($\times 6$, $\times 8$).}
    \begin{tabular}{@{\,}c@{\,}|c c c | c c c@{\,}}
    \toprule
        \rowcolor{gray!20} \multicolumn{7}{c}{\textbf{In-distribution}} \\
        \multirow{2}*{Method} & \multicolumn{3}{c}{$\times 2$} &  \multicolumn{3}{c}{$\times 4$}  \\ \cline{2-7}
        & PSNR$\uparrow$ & SSIM$\uparrow$ & SAM$\downarrow$ & PSNR$\uparrow$ & SSIM$\uparrow$ & SAM$\downarrow$ \\ \hline
        Bicubic & 29.53 & 0.599 & 0.385 & 28.11 & 0.572 & 0.396 \\
        SRNO  & 33.63 & 0.845 & 0.205 & 31.57 & 0.758 & 0.214 \\
        AnyTSR  & 34.21 & 0.857 & 0.178 & 31.71 & 0.762 & 0.204 \\
        GaussianSR  & \underline{35.98} & \underline{0.892} & \underline{0.087} & \underline{32.50} & 0.802 & \underline{0.111} \\
        SQformer  & 35.74 & 0.892 & 0.098 & 32.36 & \underline{0.803} & 0.128 \\
        MCArb  & 35.32 & 0.887 &  0.105& 32.21 & 0.795 & 0.138 \\
        Ours  & \textbf{36.28} & \textbf{0.904} & \textbf{0.075} & \textbf{32.79} & \textbf{0.820} & \textbf{0.105} \\ \hline

        \rowcolor{gray!20} \multicolumn{7}{c}{\textbf{Out-of-distribution}} \\
        \multirow{2}*{Method} & \multicolumn{3}{c}{$\times 6$} &  \multicolumn{3}{c}{$\times 8$}  \\ \cline{2-7}
        & PSNR$\uparrow$ & SSIM$\uparrow$ & SAM$\downarrow$ & PSNR$\uparrow$ & SSIM$\uparrow$ & SAM$\downarrow$ \\ \hline
        Bicubic & 27.40 & 0.564 & 0.408 & 26.89 & 0.559 & 0.418 \\
        SRNO  & 30.67 & 0.715 & 0.231 & 29.41 & 0.681 & 0.271 \\
        AnyTSR  & 30.68 & 0.713 & 0.225 & 29.53 & 0.684 & 0.268 \\
        GaussianSR  & 30.51 & 0.729 & 0.190 & 29.19 & 0.667 & 0.234  \\
        SQformer  & \underline{30.90} & \underline{0.748} & 0.187 & \underline{30.06} & \underline{0.706} & \underline{0.201} \\
        MCArb  & 30.54 & 0.733 & \underline{0.179} & 29.30 & 0.694 & 0.203 \\
        Ours  & \textbf{31.72} & \textbf{0.765} & \textbf{0.141} & \textbf{30.34} & \textbf{0.723} & \textbf{0.173} \\ 
        \bottomrule
    \end{tabular}
    \label{arbres_grss}
\end{table}
\begin{table}[!t]
    \centering
    \caption{Quantitative results of arbitrary-scale HSI super-resolution on the Chikusei dataset, including in-distribution scales ($\times 2$, $\times 4$) and out-of-distribution scales ($\times 6$, $\times 8$).}
    \begin{tabular}{@{\,}c@{\,}|c c c | c c c@{\,}}
    \toprule
        \rowcolor{gray!20} \multicolumn{7}{c}{\textbf{In-distribution}} \\
        \multirow{2}*{Method} & \multicolumn{3}{c}{$\times 2$} &  \multicolumn{3}{c}{$\times 4$}  \\ \cline{2-7}
        & PSNR$\uparrow$ & SSIM$\uparrow$ & SAM$\downarrow$ & PSNR$\uparrow$ & SSIM$\uparrow$ & SAM$\downarrow$ \\ \hline
        Bicubic & 30.26 & 0.598 & 0.306 & 29.33 & 0.592 & 0.312 \\
        SRNO  & 37.83 & 0.910 & 0.128 & 34.55 & 0.853 & 0.139 \\
        AnyTSR  & 38.06 & 0.903 & 0.126 & 34.50 & 0.844 & 0.142 \\
        GaussianSR  & 42.03 & 0.969 & \underline{0.047} & \underline{36.36} & 0.899 & 0.068   \\
        SQformer  & \underline{42.12} & \underline{0.970} & 0.049 & 36.32 & \underline{0.901} & \underline{0.067} \\
        MCArb  & 41.25 & 0.944 & 0.067 & 35.03 & 0.883 & 0.093 \\
        Ours  & \textbf{42.37} & \textbf{0.971} & \textbf{0.045} & \textbf{36.42} & \textbf{0.904} & \textbf{0.065} \\ \hline

        \rowcolor{gray!20} \multicolumn{7}{c}{\textbf{Out-of-distribution}} \\
        \multirow{2}*{Method} & \multicolumn{3}{c}{$\times 6$} &  \multicolumn{3}{c}{$\times 8$}  \\ \cline{2-7}
        & PSNR$\uparrow$ & SSIM$\uparrow$ & SAM$\downarrow$ & PSNR$\uparrow$ & SSIM$\uparrow$ & SAM$\downarrow$ \\ \hline
        Bicubic & 28.70 & 0.590 & 0.317 & 28.20 & 0.582 & 0.320 \\
        SRNO  & 32.64 & 0.804 & 0.150 & 31.40 & 0.763 & 0.164 \\
        AnyTSR  & 32.30 & 0.787 & 0.163 & 31.47 & 0.755 & 0.168 \\
        GaussianSR  & 33.35 & 0.834 & 0.102 & 31.94 & 0.776 & 0.125 \\
        SQformer  & \underline{33.87} & \underline{0.852} & \underline{0.086} & \underline{32.19} & \underline{0.797} & \underline{0.106} \\
        MCArb  & 32.39 & 0.822 & 0.124 & 31.31 & 0.766 & 0.126 \\
        Ours  & \textbf{33.93} & \textbf{0.857} & \textbf{0.084} & \textbf{32.24} & \textbf{0.801} & \textbf{0.104} \\ 
        \bottomrule
    \end{tabular}
    \label{arbres_chi}
\end{table}

\subsection{Arbitrary-scale HSI super-resolution}
For arbitrary-scale HSI super-resolution, we compare our GaussianHSI with SQformer~\cite{sqformer} and MCArb~\cite{mcarb}. Given the limited number of methods specifically developed for this task, we also include several arbitrary-scale super-resolution methods originally proposed for other image modalities, including SRNO~\cite{srno}, AnyTSR~\cite{anytsr}, and GaussianSR~\cite{gaussiansr}. All methods are trained using scale factors ranging from $\times2$ to $\times4$ and evaluated at $\times2$, $\times4$, $\times6$, and $\times8$. This setting evaluates reconstruction performance both within and beyond the training scale range, following SQformer~\cite{sqformer}. Gaussian noise with a noise level of 10 is added to the low-resolution HSIs to evaluate performance under challenging noisy conditions~\cite{hirdiff}. All methods are carefully tuned for this setting and evaluated using the same data and evaluation protocol to ensure a fair comparison.

\begin{figure*}[!t]
    \centering
    \includegraphics[width=0.95\linewidth]{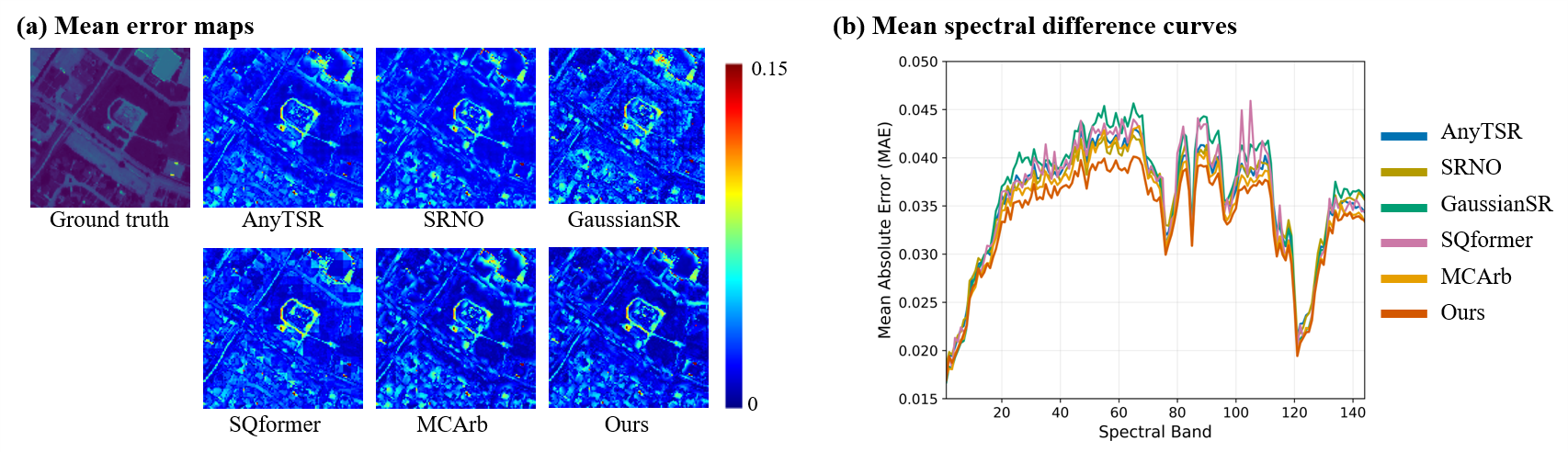}
    \caption{Comparison of reconstruction errors for different methods in arbitrary-scale HSI super-resolution on the Houston dataset at scale $\times$6. (a) Ground-truth image and spatial error maps obtained by averaging absolute errors over spectral bands. (b) Band-wise mean absolute error curves averaged over spatial pixels.}
    \label{fig:error_arb}
\end{figure*}

Quantitative results on Pavia, Houston, and Chikusei are reported in TABLEs~\ref{arbres_pc}, \ref{arbres_grss}, and \ref{arbres_chi}, respectively. Since all models are trained on scales from $\times2$ to $\times4$, evaluation at scales $\times2$ and $\times4$ is considered in-distribution with respect to scale, as these factors lie within the training range. Evaluation at scales $\times6$ and $\times8$ is considered out-of-distribution and assesses generalization to unseen scale factors beyond the training range. GaussianHSI achieves the best results in PSNR, SSIM, and SAM across all three datasets at the evaluated in-distribution and out-of-distribution scales. These results indicate improvements in spatial reconstruction quality and spectral fidelity over the compared methods. Compared with SQformer and MCArb, which are developed for arbitrary-scale HSI super-resolution, GaussianHSI achieves consistent improvements across the evaluated settings. For example, on Pavia, GaussianHSI improves PSNR over SQformer by 1.39~dB and 0.51~dB at scales $\times2$ and $\times8$, respectively. On Houston at the unseen scale $\times6$, it increases PSNR from 30.90~dB to 31.72~dB and reduces SAM from 0.187 to 0.141 relative to SQformer. In comparison, several arbitrary-scale methods originally developed for other image modalities show relatively limited super-resolution performance, particularly in terms of SAM. This gap may be related to differences in spatial and spectral characteristics between the original image modalities and remote sensing HSIs. Among these methods, GaussianSR also adopts a continuous Gaussian representation and achieves competitive results. GaussianHSI achieves better reconstruction performance than GaussianSR. These results suggest that the adaptive spatial aggregation of VBGS and the spectral refinement of SDE can complement Gaussian representations to improve HSI super-resolution performance.

Visual comparisons on the Houston dataset are presented in Fig.~\ref{fig:arbres} at scales $\times2$ and $\times6$. GaussianHSI reconstructs clearer spatial structures and produces local details that are visually closer to the ground truth, particularly at the unseen scale $\times6$. These observations are consistent with the quantitative improvements in PSNR and SSIM. Fig.~\ref{fig:error_arb} further compares reconstruction errors at scale $\times6$ through spatial maps of absolute errors averaged over spectral bands and band-wise mean absolute error curves averaged over spatial pixels. GaussianHSI exhibits lower reconstruction errors in the spatial maps and across most spectral bands. Overall, the quantitative and qualitative results demonstrate the effectiveness of GaussianHSI for arbitrary-scale HSI super-resolution.

\subsection{Model Analysis}
In this section, we report analysis results on the Houston dataset at scale $\times 2$. Similar results can be observed on the other datasets and settings. 

\begin{table}[!t]
    \centering
    \caption{Efficiency analysis on the Houston dataset at scale $\times 2$ for inference on a single low-resolution HSI.}
    \begin{tabular}{c| c c c c}
    \toprule
        & MCNet & MSDformer & MCArb & ESSAformer \\ \hline
        Params & 1.93M & 12.63M & 8.95M & 9.21M \\
        FLOPs  & 5.21T & 118.34G & 367.97G & 187.69G \\ \hline
        & HDL & HSRMamba & SQformer & \textbf{Ours} \\ \hline
        Params & 404.75K & 937.17K & 19.50M & 1.27M \\
        FLOPs  & 5.84G & 5.878G & 74.41G & 9.76G \\ 
    \bottomrule
    \end{tabular}
    \label{tab:efficiency}
\end{table}
\noindent \textbf{Computational Efficiency.} TABLE~\ref{tab:efficiency} reports the parameter counts (Params) and floating-point operations (FLOPs) of the compared methods for inference on a single low-resolution HSI from the Houston dataset at scale $\times2$. GaussianHSI contains 1.27M parameters and requires 9.76G FLOPs, indicating competitive model complexity and computational cost. Compared with SQformer, which achieves the second-best reconstruction results in this setting, GaussianHSI achieves better reconstruction quality with substantially fewer parameters and FLOPs. GaussianHSI also requires fewer parameters and FLOPs than MCArb, another method designed for arbitrary-scale HSI super-resolution. Furthermore, once the Gaussian parameters have been predicted, Gaussian selection, weighting, and aggregation can be performed independently for different target high-resolution pixels. These operations can therefore be performed simultaneously for multiple target pixels, allowing parallel GPU computation during inference.

\begin{figure}[!t]
    \centering
    \includegraphics[width=0.73\linewidth]{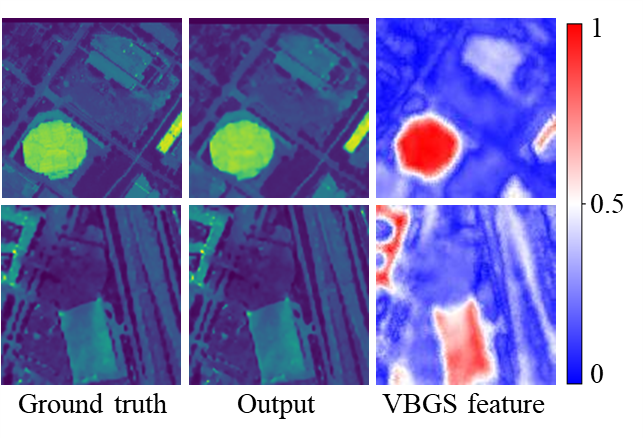}
    \caption{Visualization of spatial features from the VBGS. The maximum value across all channels is used for visualization.}
    \label{fig:fea}
\end{figure}

\noindent \textbf{Effectiveness of VBGS.} 
We conduct an ablation study to evaluate the contribution of the proposed Voronoi-guided Bilateral 2D Gaussian Splatting (VBGS) module. Specifically, VBGS is replaced with the Gaussian splatting module adopted in GaussianSR~\cite{gaussiansr}. As shown in TABLE~\ref{tab:abla}, this replacement leads to degraded super-resolution performance, indicating the effectiveness of VBGS. Remote sensing HSIs contain structures with varying spatial extents and shapes. VBGS jointly considers spatial relevance and feature consistency with the low-resolution reference to adapt Gaussian aggregation to these structural variations, which may help recover spatial details. Fig.~\ref{fig:fea} further presents visualizations of the spatial features reconstructed by VBGS. The feature maps exhibit responses to image structures, illustrating the spatial information retained in the reconstructed representations. These ablation results and feature visualizations indicate the contribution of VBGS to spatial reconstruction in GaussianHSI.

\begin{figure*}[!t]
    \centering
    \includegraphics[width=1\linewidth]{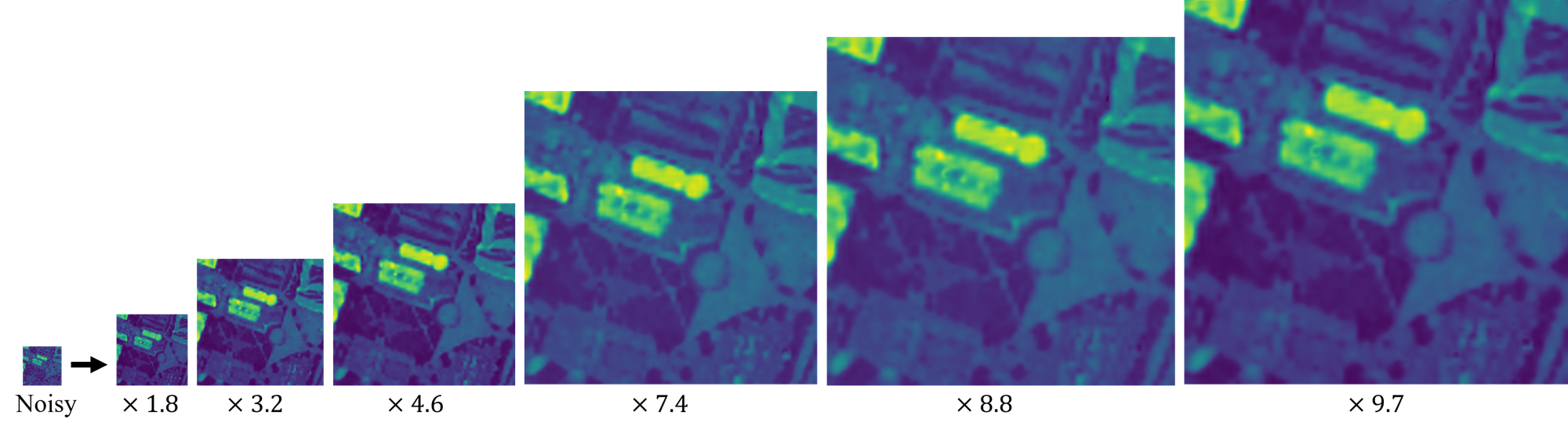}
    \caption{Visual results of HSI super-resolution at non-integer scale factors on the Houston dataset using a synthetically downsampled input corrupted with Gaussian noise. The 23rd spectral band is visualized.}
    \label{fig:float_res}
\end{figure*}

\begin{table}[!t]
    \centering
    \caption{Ablation study of the VBGS and SDE modules.}
    \begin{tabular}{c c|c c c}
    \toprule
        \textbf{VBGS} & \textbf{SDE}  & PSNR & SSIM & SAM \\ \hline
        \ding{55}  & \ding{51}  & 36.87 & 0.923 & 0.070 \\ 
        \ding{51} & \ding{55}  & 36.82 & 0.919 & 0.072 \\ 
         \ding{51}  & \ding{51}  & 37.12 & 0.929 & 0.069 \\ 
    \bottomrule
    \end{tabular}
    \label{tab:abla}
\end{table}

\noindent \textbf{Effectiveness of SDE.}
We conduct an ablation study to evaluate the contribution of the proposed Spectral Detail Enhancement (SDE) module. In the variant without SDE, the module is bypassed. Its input features are directly passed to the subsequent reconstruction process. The remaining network is unchanged. As shown in TABLE~\ref{tab:abla}, removing SDE leads to degraded reconstruction performance. In particular, the higher SAM value indicates greater spectral distortion in the reconstructed high-resolution HSIs. SDE incorporates local information to refine spectral representations. This refinement may help recover spectral details and complement the spatial reconstruction performed by VBGS. Overall, the ablation results demonstrate the contribution of SDE to spectral fidelity and overall reconstruction quality in GaussianHSI.

\noindent \textbf{Impact of Top-$k$ selection.} 
We conduct a parameter sensitivity analysis to evaluate the effect of the number of selected Gaussian functions, $k$, in VBGS. TABLE~\ref{tab:topk} reports the reconstruction performance under different values of $k$. PSNR improves slightly as $k$ increases from 8 to 16 and remains similar at $k=32$. When $k$ is small, spatial information relevant to reconstruction may not be fully utilized, which could lead to degraded super-resolution performance. As $k$ increases further, PSNR gradually decreases. A larger candidate set may include additional Gaussian functions with lower relevance to the target pixel~\cite{continuoussr}. Aggregating these additional contributions may also increase the difficulty of optimization~\cite{3dgs}. These factors provide possible explanations for the observed trend. The results suggest that an appropriate choice of $k$ is important for reconstruction quality. Based on this sensitivity analysis, we set $k=16$ in our experiments.

\begin{table}[!t]
    \centering
    \caption{Parameter sensitivity analysis of $k$ in the VBGS.}
    \begin{tabular}{c | c c c c c c c}
    \toprule
        \textbf{top-}$k$ & 8 & 16 & 32 & 64 & 128 & 512 & 1024 \\ \hline
        PSNR  & 37.09 & 37.12 & 37.11 & 37.09 & 37.05 & 36.84 & 36.72 \\ 
    \bottomrule
    \end{tabular}
    \label{tab:topk}
\end{table}

\noindent \textbf{Impact of local patch size.}
We conduct a parameter sensitivity analysis to evaluate the effect of patch size in the SDE module. TABLE~\ref{tab:patch} reports the reconstruction performance under different patch sizes. A patch size of 3 yields a PSNR of 36.98 dB. Increasing the patch size to 5 improves PSNR to 37.12 dB. This improvement suggests that local spatial context provides useful complementary information for spectral detail enhancement. With a smaller patch, relevant neighboring information may not be fully utilized, which may lead to lower reconstruction performance. Further increases in patch size continue to improve PSNR, but the gains become progressively smaller. Specifically, increasing the patch size from 5 to 11 yields an additional improvement of 0.08 dB. This trend suggests that reconstruction performance approaches saturation as the patch size increases. A patch size of 5 already achieves reconstruction performance close to the performance obtained with larger patches. Based on this sensitivity analysis, we set the patch size to 5 in all experiments.

\begin{table}[!t]
    \centering
    \caption{Parameter sensitivity analysis of patch size in the SDE module.}
    \begin{tabular}{c|c c c c c}
    \toprule
        \textbf{Patch size} & 3 & 5 & 7 & 9 & 11 \\ \hline
        \textbf{PSNR} & 36.98 & 37.12 & 37.17 & 37.19 & 37.20 \\
    \bottomrule
    \end{tabular}
    \label{tab:patch}
\end{table}

\noindent \textbf{Super-resolution at Non-integer Scales.}
We further evaluate GaussianHSI at a series of non-integer scale factors on the Houston dataset, including in-distribution scales within the training range and out-of-distribution scales beyond this range. The input HSI is downsampled by a factor of two and then corrupted with Gaussian noise at a noise level of 10. The resulting low-resolution observation is then reconstructed at different target scales using a model trained with the same strategy as in the arbitrary-scale experiments. No additional training is performed for the evaluated scales. As shown in Fig.~\ref{fig:float_res}, the reconstructed images preserve the spatial structures and fine-grained details. Major structures remain discernible even at scales beyond the training range, such as $\times9.7$. These qualitative results illustrate the flexibility of GaussianHSI in performing super-resolution at non-integer scales.

\section{Conclusion}
\label{sec5}

In this paper, we proposed GaussianHSI for arbitrary-scale HSI super-resolution. The proposed method combines a Voronoi-Guided Bilateral 2D Gaussian Splatting module for adaptive spatial reconstruction with a Spectral Detail Enhancement module for complementary spectral recovery. By introducing adaptive Voronoi-guided Gaussian selection and reference-aware bilateral aggregation, GaussianHSI improves the flexibility of Gaussian-based reconstruction under varying spatial structures. Extensive experiments on benchmark datasets demonstrate its effectiveness in both single-scale and arbitrary-scale HSI super-resolution. We hope this work can provide useful insights for future research on Gaussian-based HSI restoration. Future work will examine the potential of extending this framework to image restoration tasks in other domains.

\bibliographystyle{IEEEtran}
\bibliography{ref}

\newpage

\end{document}